\documentclass[10pt,twocolumn,letterpaper]{article}

\usepackage{cvpr}              

\usepackage{graphicx}
\usepackage{amsmath}
\usepackage{amssymb}
\usepackage{booktabs}
\usepackage{multirow}

\usepackage[pagebackref,breaklinks,colorlinks]{hyperref}

\usepackage[capitalize]{cleveref}
\crefname{section}{Sec.}{Secs.}
\Crefname{section}{Section}{Sections}
\Crefname{table}{Table}{Tables}
\crefname{table}{Tab.}{Tabs.}

\def\cvprPaperID{36} 
\def\confName{CVPR}
\def\confYear{2022}

\begin{document}

\title{Multiple Degradation and Reconstruction Network for Single Image Denoising via Knowledge Distillation}

\author{Juncheng Li$^{1,2}$,\,Hanhui Yang$^1$,\,Qiaosi Yi$^2$,\,Faming Fang$^2$,\,Guangwei Gao$^3$,\,Tieyong Zeng$^1\thanks{Corresponding author}$,\,Guixu Zhang$^2$\\
$^{1}$The Chinese University of Hong Kong
$^{2}$East China Normal University\\
$^{3}$Nanjing University of Posts and Telecommunications\\
{\tt\small \{cvjunchengli, qiaosiyijoyies, csggao\}@gmail.com}\\
{\tt\small \{hhYang, zeng@math\}@math.cuhk.edu.hk, \{fmfang, gxzhang\}@cs.ecnu.edu.cn}
}

\maketitle

\begin{abstract}
Single image denoising (SID) has achieved significant breakthroughs with the development of deep learning.
However, the proposed methods are often accompanied by plenty of parameters, which greatly limits their application scenarios.
Different from previous works that blindly increase the depth of the network, we explore the degradation mechanism of the noisy image and propose a lightweight Multiple Degradation and Reconstruction Network (MDRN) to progressively remove noise.
Meanwhile, we propose two novel Heterogeneous Knowledge Distillation Strategies (HMDS) to enable MDRN to learn richer and more accurate features from heterogeneous models, which make it possible to reconstruct higher-quality denoised images under extreme conditions.
Extensive experiments show that our MDRN achieves favorable performance against other SID models with fewer parameters. Meanwhile, plenty of ablation studies demonstrate that the introduced HMDS can improve the performance of tiny models or the model under high noise levels, which is extremely useful for related applications.
\end{abstract}

\section{Introduction}
Single image denoising (SID) aims to reconstruct a clean image from the noisy one, which has been widely used as an initial step for many high-level computer vision tasks.
This is because the quality of the denoised images will significantly influence the accuracy of these downstream tasks.
However, it still is a challenging task since many important features in the original image has be seriously occluded or corrupted by noise.

\begin{figure}
  \centering
  \begin{minipage}[c]{1\textwidth}
  \includegraphics[width=8cm]{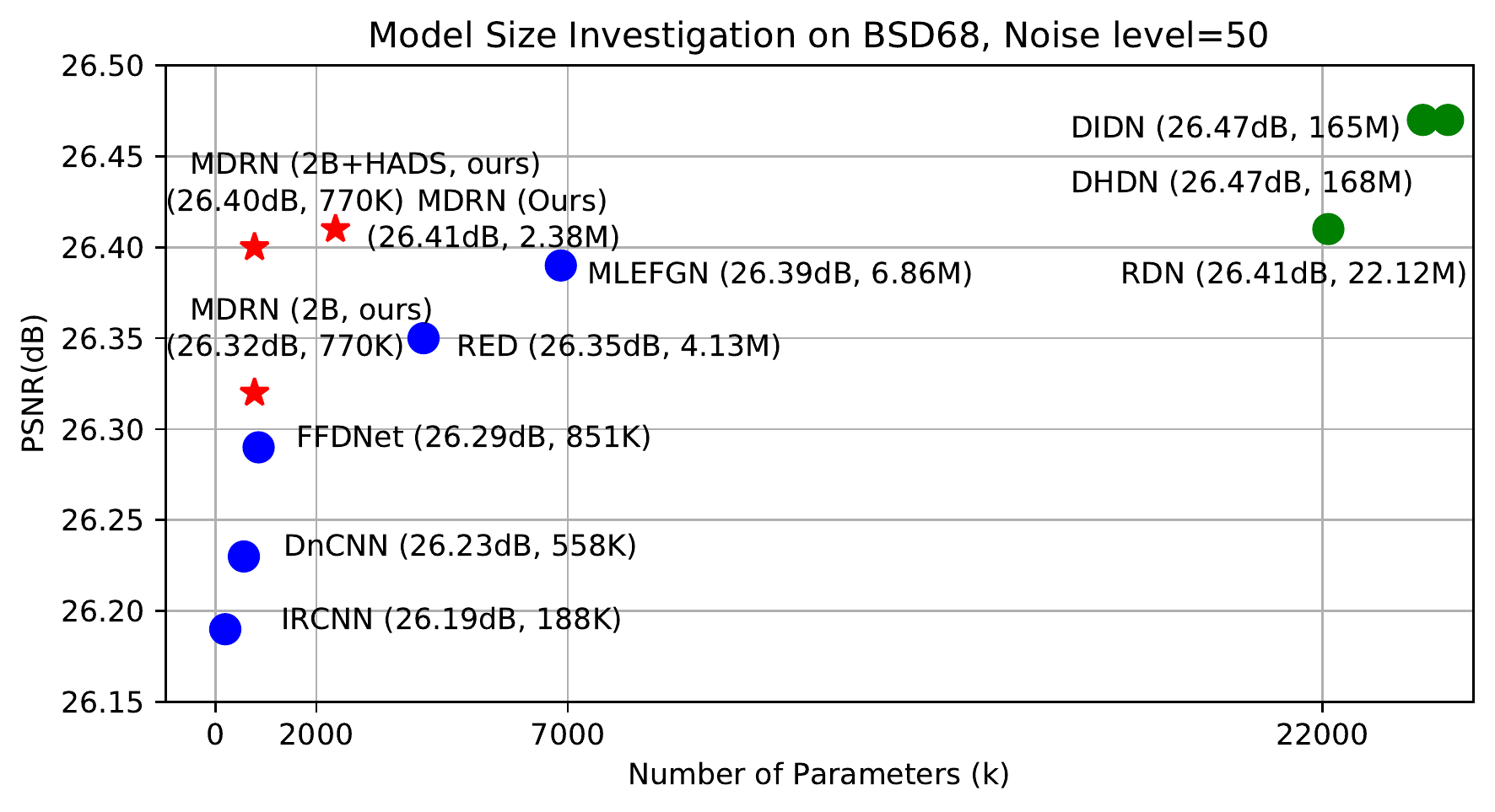}
  \end{minipage}
  \caption{Model size investigation. Red stars denote our models.}
  \label{parameter}
\end{figure}

\begin{figure*}
\centering
\includegraphics[width=0.8\textwidth]{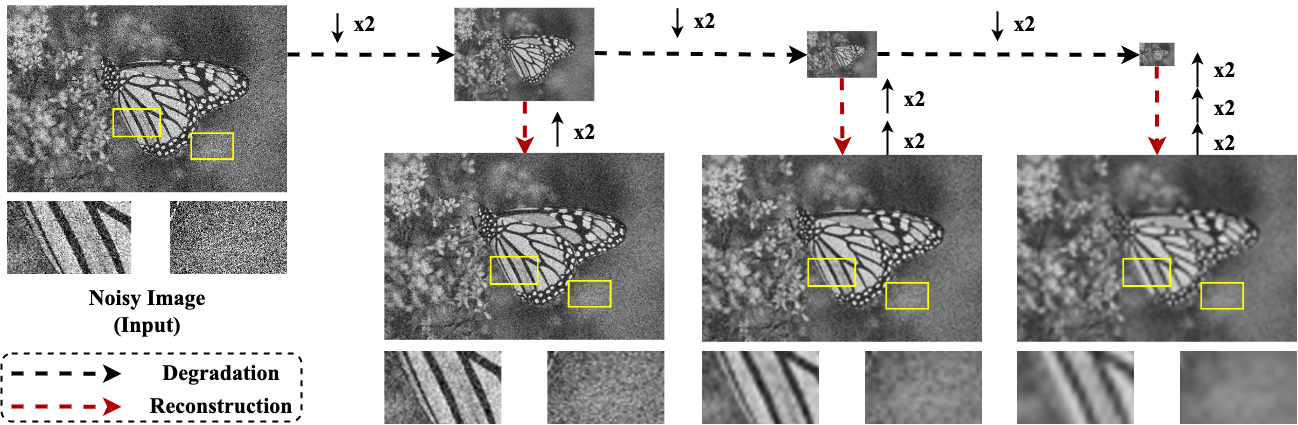}
\caption{An example of image denoising ($\sigma=50$) by using multiple degradation and reconstruction operations (via Bicubic).}
\label{Motivation}
\end{figure*}

During the past decades, the Convolutional Neural Network (CNN) has achieved remarkable achievements in image processing~\cite{dong2015image, nah2017deep, yi2021efficient}, which also greatly promotes the development of SID.
Recently, many CNN-based models have been proposed for SID, especially for the Additive White Gaussian Noise (AWGN).
Different from traditional methods, these methods~\cite{zhang2017beyond,yang2018bm3d,liu2019multi,fang2020multilevel,Park2019DenselyCH,Yu2019DeepID,lan2021image,sharif2020learning} usually learn the mapping between noisy and clear images by building a well-designed CNN.
For example, Zhang et al.~\cite{zhang2017beyond} proposed the first end-to-end trainable CNN model (DnCNN) for Gaussian denoising, which took the advantage of batch normalization and residual learning to recover clean images.
Meanwhile, Zhang et al.~\cite{zhang2018ffdnet} also proposed a flexible FFDNet, which took the noisy image and estimated noise level map as inputs, thus a single model could deal with noise on different levels.
Liu et al.~\cite{liu2019multi} presented a multi-level wavelet CNN (WCNN) for better trade-off between the receptive field size and the computational efficiency.
Fang et al.~\cite{fang2020multilevel} introduced image edge prior into CNN and proposed a Multilevel Edge Features Guided Network (MLEFGN) for SID.
On the other hand, some large models have also been proposed and constantly refreshed the best results.
For instance, Zhang et al.~\cite{zhang2020residual} proposed a Residual Dense Network (RDN) for image restoration with residual dense block.
Park et al.~\cite{Park2019DenselyCH} studied a Densely Connected Hierarchical Network (DHDN) by using a modified U-net architecture;
Although the above models achieve promising results, they still do not handle high-noise images well.
Meanwhile, these models are often accompanied by a large number of parameters (Figure \ref{parameter}), which greatly limits their application in real scenarios.

Considering the popularity of smart mobile devices, it is extremely important to build a lightweight and efficient SID model.
As shown in Figure~\ref{Motivation}, we find that some noise in the noisy image can be eliminated by using the simple down- and up-sampling operations.
However, we also notice that the downsampling operation also results in the loss of useful information when removing noise.
Meanwhile, the bigger the downsampling rate, the more noise will be removed, but the reconstructed image will become more blurred.
To solve the aforementioned problem, we propose a Multiple Degradation and Reconstruction Network (MDRN) to progressively remove image noise inspired by Unet++ (\cite{zhou2018unet}).
Meanwhile, Multi-scale Aggregation Block (MSAB) and Multi-scale Aggregation Group (MSAG) are specially designed for multi-scale feature extraction.
Meanwhile, we propose two novel Heterogeneous Knowledge Distillation Strategies (HKDS) to enable MDRN to learn richer and more accurate features from the teacher model, thereby further improving the model performance.
In other words, the model uses the knowledge provided by external heterogeneous models to realize the automatic learning of deep feature priors. In summary, the contributions of this paper are as follows:

(1). We propose a lightweight and efficient Multiple Degradation and Reconstruction Network (MDRN) for image denoising, which can progressively remove image noise via multiple down- and up-sampling operations.

(2). We propose a Multi-Scale Aggregation Group (MSAG) for feature extraction. MSAB is the basic components of MDRN, which can extract rich multi-scale features with few parameters.

(3). We propose two Heterogeneous Knowledge Distillation Strategies (HKDS) for SID. With the help of HKDS, MDRN can learn richer and more accurate features from the teacher model, thereby improving the denoising ability of the lightweight model under extreme conditions.

\section{Related Work}
\subsection{Image Denoising}
The degradation mode of the noisy image is usually formulated as $y = x + v$, where $x$ is the clean image, $v$ is the additive white Gaussian noise (AWGN), and $y$ is noisy image polluted by AWGN.
Recently, many excellent methods have been proposed for clear images estimation, including spatial filtering methods~\cite{buades2005non,he2013guided}, model-based methods~\cite{dong2013nonlocally,tansey2018maximum}, and deep learning-based methods~\cite{zhang2017beyond,zhang2018ffdnet,yang2018bm3d,liu2019multi,fang2020multilevel,zhang2020residual,zhang2021drnet,lan2021image,sharif2020learning}.
Among them, deep learning-based methods have become the mainstream of SID, which aim to solve the problem via learning the mapping between the noisy and clean images.
For example, Zhang et al.~\cite{zhang2017beyond} proposed a DnCNN for the Gaussian noise removal, which achieved competitive results by took advantage of batch normalization and residual learning.
Yu et al.~\cite{Yu2019DeepID} proposed a Deep Iterative Down-up Network (DIDN) for image denoising, which achieved promising results.
However, directly minimizing the loss function is difficult to learn accurate mapping.
To solve this problem, Fang et al.~\cite{fang2020multilevel} introduced the edge priors to guide image denoising and proposed a Multi-level Edge Features Guided Network (MLEFGN).
However, the edge prior still is an artificially designed image prior and it is only one of many image priors. 
This certain extent limits its flexibility and versatility so that it can only improve the local effects in the image. 
Different from it, we aim to explore a method that allows the model to automatically learn the deep priors according to the need of the image, and to build a more efficient and lightweight SID model.

\begin{figure*}
\centering
\includegraphics[width=0.76\textwidth]{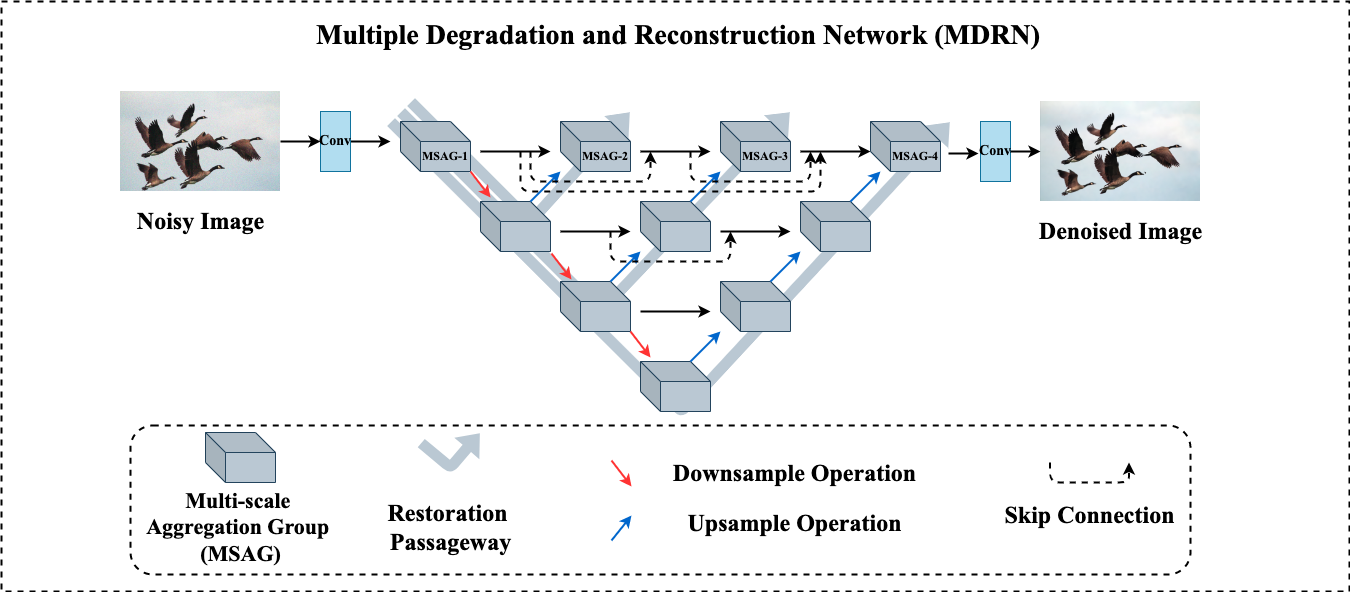}
\caption{The architecture of Multiple Degradation and Reconstruction Network (MDRN). It is worth noting that convolutional and deconvolutional layers are introduced to replace the original down and up-sampling operations performed by Bicubic.}
\label{MDRN}
\end{figure*}

\subsection{Knowledge Distillation}
Knowledge distillation (KD) was first proposed by Hinton et al.~\cite{hinton2015distilling}, also known as the teacher-student framework.
The goal of the KD mechanism is to transfer the knowledge from a complex model (Teacher) to a simple one (Student), which is often used to compress model size.
Recently, many different KD strategies have been proposed, including output transfer~\cite{hinton2015distilling, zhang2018deep}, feature transfer~\cite{romero2014fitnets,zagoruyko2016paying}, and relation transfer~\cite{yim2017gift}.
All these mechanisms have been widely used in the image classification task and achieved remarkable results.
Meanwhile, some researchers are also trying to introduce the KD mechanism into the image restoration tasks.
For example, Hong et al.~\cite{hong2020distilling} proposed a KD-based method for image dehazing with the heterogeneous task.
Specifically, the teacher model is an off-the-shelf autoencoder network, used to learn the reconstruction process from clear to clear images and transfer the learned knowledge to the dehazing model during the training process.
Learning knowledge from clear images is an interesting idea.
However, our investigation shows that the reconstruction of clear to clear images using a flat network such as U-net~\cite{ronneberger2015u}, is a relatively simple task that is difficult to ensure all layers or modules in the teacher model are fully trained since a flat network with skip connection could directly transmit all original information to the last layer. Therefore, it will make the teacher model fail to guide the student model efficiently.
In this paper, we aim to explore more interpretable knowledge distillation strategies for SID to further improve the denoising ability of the tiny model, especially under high noise levels.

\section{Proposed Method}
In this paper, we propose a lightweight and efficient Multiple Degradation and Reconstruction Network (MDRN) for SID.
Meanwhile, we present a Heterogeneous Architecture Distillation Strategy (HADS) and a Heterogeneous Mode Distillation Strategy (HMDS) to automatically learn the required deep prior knowledge from heterogeneous models, thereby further improving the model performance.

\subsection{MDRN}
As shown in Figure~\ref{MDRN}, the core part of the Multiple Degradation and Reconstruction Network (MDRN) is an inverted pyramid, which consists of a series of Multi-scale Aggregation Groups (MSAGs).
Meanwhile, we can clearly observe that the model contains multiple degradation and reconstruction operations, which are used to remove noise and restore high-frequency detail, respectively.
It is worth noting that convolutional and deconvolutional layers are introduced to replace the original down and up-sampling operations performed by Bicubic.
This method can automatically learn how to remove useless noises while preserving useful information.
In addition, after each downsampling operation, we will restore it immediately, thus we can obtain rich features reconstructed after different scales of downsampling operations.
Furthermore, we introduce the dense connection strategy at each level of the model to improve the utilization of hierarchical features, thus further improving the model performance.

\subsubsection{Multi-scale Aggregation Group}
\begin{figure}
\centering
\includegraphics[width=6.3cm]{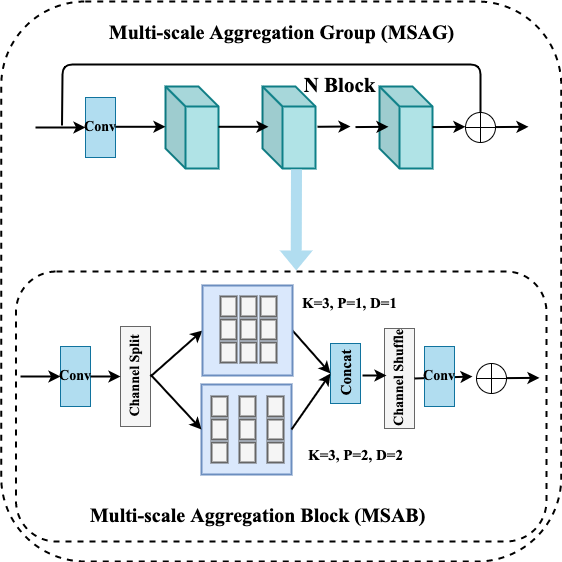}
\caption{The architecture of MSAG. K, P, and D represent the kernel size, padding, and dilation rate, respectively.}
\label{MSAG}
\end{figure}

The Multi-scale Aggregation Group (MSAG) is the most basic and important module of MDRN.
According to Figure~\ref{MSAG}, we can observe that MSAG is composed of a $1 \times 1$ fusion layer and $N$ MSABs, which is a simple but efficient feature extraction module.
Meanwhile, we also apply a long skip connection for global residual learning like~\cite{zhang2018image}.
This strategy can increase the information flow and solve the gradient disappearance problem.

\textbf{Multi-scale Aggregation Block.} In order to extract rich features, we propose a Multi-scale Aggregation Block (MSAB). 
MSAB is an efficient feature extraction module that can extract rich multi-scale features with few parameters, which is inspired by MSRB~\cite{li2018multi} and MDCB~\cite{li2020mdcn}, and the number of parameters of the module is reduced by using the channel splitting mechanism and dilated convolutional layers.
As shown in Figure~\ref{MSAG}, after a $1 \times 1$ convolutional layer, we use the channel splitting operation to divide the features into two groups.
Then, we apply the convolutional layer with different dilation rates on different groups to extract image features with different scales.
After that, all extracted features are concatenated, and the channel shuffle operation is applied to overcome the side effects brought by the channel splitting.
Finally, a $1 \times 1$ convolutional layer is used for feature fusion and the residual learning strategy is introduced to further improve the information flow.

\begin{figure*}
\centering
\includegraphics[width=0.8\textwidth]{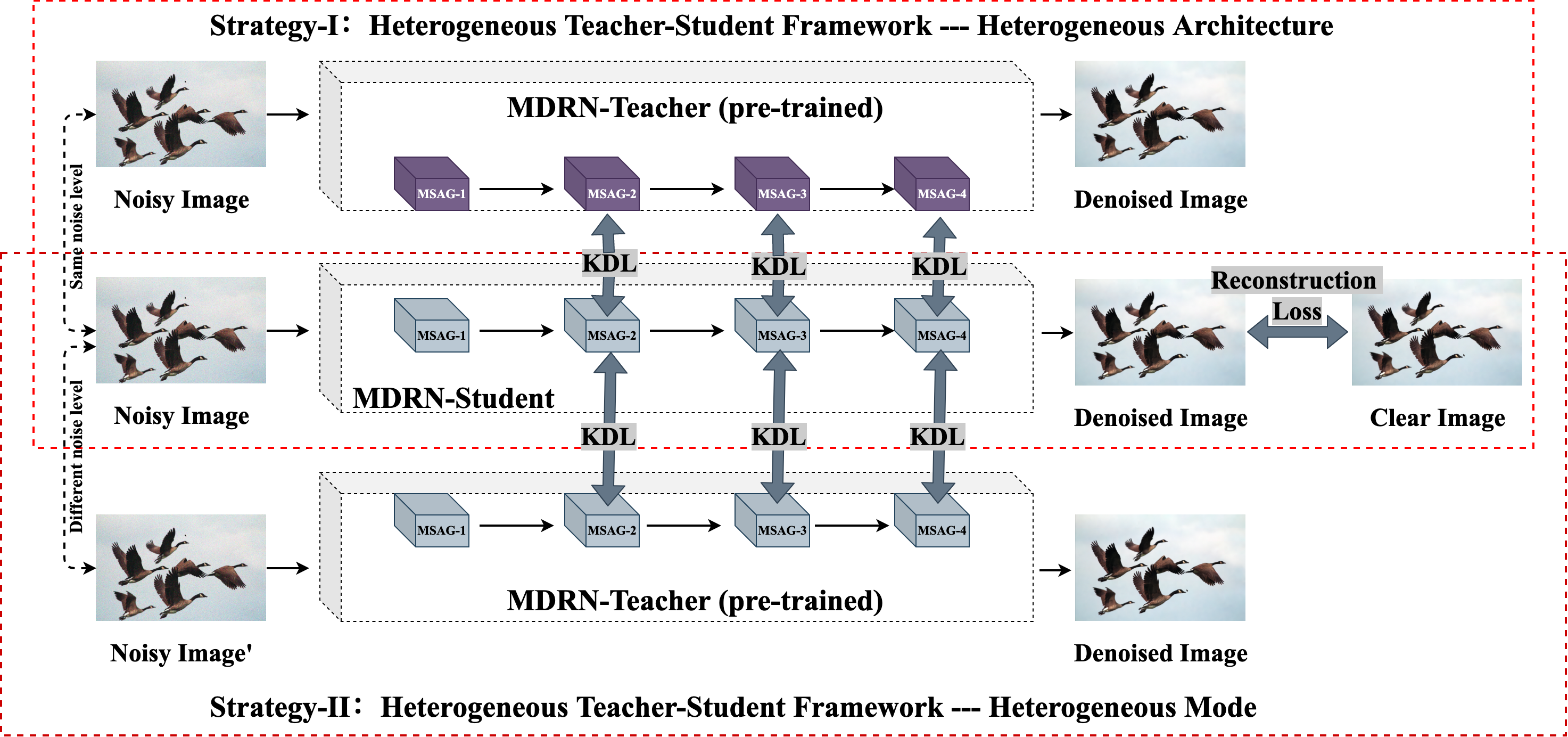}
\caption{Schematic diagram of two heterogeneous knowledge distillation strategies. It is worth noting that only the top 4 MSAGs of MDRN are visualized in this figure for a clearer illustration	.}
\label{KDS}
\end{figure*}

\textbf{Hierarchical Dense Connection.}
MDRN is a multiple degradation and reconstruction model, thus can produce rich degraded and restored features.
However, if these features are independent of each other, it is not beneficial for the final denoised image reconstruction.
To solve this problem, we introduce dense connections at each level to fully utilize hierarchical features.
Specifically, the current MSAG receives the feature maps of all preceding MSAGs $x^{j}_{0}, x^{j}_{1}, \cdots,$ and $x^{j}_{i-1}$ in the same level as inputs
\begin{equation}
  x^{j}_{i} = H^{j}_{i}([x^{j}_{0}, x^{j}_{1}, \cdots, x^{j}_{i-1}]),
\end{equation}
where $[x^{j}_{0}, x^{j}_{1}, \cdots, x^{j}_{i-1}]$ refers to the concatenation operation and $H^{j}_{i}(\cdot)$ denotes the $i^{th}$ MSAG in the $j^{th}$ level.

\subsection{Heterogeneous Knowledge Distillation}
Although MDRN is an effective SID model, it is still difficult to reconstruct high-quality denoised images under extreme conditions, such as tiny models or severe noises.
Therefore, we propose two novel Heterogeneous Knowledge Distillation Strategies (HKDS) to enable MDRN to automatically learn the deep feature priors from the teacher model.
Different from previous works~\cite{hong2020distilling, wu2020knowledge, li2020pams} that learned the knowledge from different models or tasks, we explore more efficient knowledge distillation strategies from the model and the task itself.

\textbf{Heterogeneous Architecture Distillation (HADS).}
HADS aims to transfer the knowledge from a large model to the small one.
As shown in Figure~\ref{KDS} (Top), the framework contains two models, a large teacher model and a student model.
\textbf{Both of these two models have the same backbone, but the number of MSAB in each MSAG of these two models is different.}
In other words, the purple module has more MSAB than the gray one.
Except for this, all settings of these two models are consistent, including the noise level.
Under this setting, each MSAG in the teacher model can extract richer and more accurate features than the student.
We hope that the features extracted by MSAG in the student model can be as similar as possible to the features extracted by MSAG in the teacher model.
To achieve this, we introduce Knowledge Distillation Loss (KDL) between the reconstructed outputs (MSAG-2, MSAG-3, and MSAG-4) of these two models to enable the student model can learn more accurate features from the teacher.
With the help of the HADS, the student model can converge faster and the model performance can be further improved, which benefits for lightweight and accurate model construction.

\textbf{Heterogeneous Mode Distillation (HMDS).}
Although the deep model can learn more accurate features, it is still difficult to deal with severe noise (e.g., $\sigma=70$) since the input image has been seriously damaged.
In this case, changing the size of the model itself is difficult to effectively improve the model performance.
To address this issue, we propose a Heterogeneous Mode Distillation Strategy (HMDS).
As shown in Figure~\ref{KDS} (Bottom), the framework also contains two models.
It is worth noting that these two models are exactly the same in model structure and size.
The only difference is that the noise levels handled by these two models are different.
In other words, HMDS aims to transfer the knowledge from a denoising model designed for slight noise to a model designed for severe noise.
For example, the denoising model designed for low noise levels (e.g., $\sigma=30$, $\sigma=50$) is used as the teacher model to guide the model designed for high noise levels (e.g., $\sigma=70$).
This is because low noise level images are easier to restored than high noise level images.
Therefore, the student model can learn from the teacher model more accurate high-frequency details which are absent from its own input image.
Meanwhile, the model can recover more accurate texture details even under high noise levels.
The same as the HADS, we also introduce knowledge distillation loss between these two models.

\subsection{Loss Function}
\textbf{Reconstruction Loss.} Following previous works, we also adopt L1 loss to measure the difference between the denoised image and the corresponding clean image
\begin{equation}
   \mathcal{L}_{RL} = \left \| F_{SID}(I_{noisy}) - I_{clear} \right \| _{1},
\end{equation}
where $I_{noisy}$ and $I_{clear}$ represent the noisy input and corresponding clear image, respectively.
$F_{SID}(\cdot)$ denotes the proposed MDRN and $F_{SID}(I_{noisy})$ is the denoised result.

\begin{table*}
   \centering
   \small
  \setlength{\tabcolsep}{3.3mm}
  \renewcommand{\arraystretch}{0.85}
   \begin{tabular}{r|c|c|c|c|c|c|c|c|c}
   \hline
   Method   & \multicolumn{3}{c|}{Set12}    & \multicolumn{3}{c|}{BSD68}     & \multicolumn{3}{c}{Urban100} \\ \hline
   Noise Level   & $\sigma$=15      & $\sigma$=25      & $\sigma$=50     & $\sigma$=15       & $\sigma$=25      & $\sigma$=50       & $\sigma$=15    & $\sigma$=25    & $\sigma$=50    \\ \hline 
   BM3D        & 32.37   & 29.97   & 26.72     & 31.08     & 28.57    & 25.62    & 32.34 & 29.70 & 25.94   \\
   RED30          & 32.83   & 30.48   & 27.34     & 31.72     & 29.26    & 26.35    & 32.75 & 30.21 & 26.64 \\
   TNRD      & 32.50   & 30.06   & 26.81     & 31.42     & 28.92    & 25.97    & 31.98 & 29.29 & 25.71   \\
   IRCNN   & 32.77   & 30.38   & 27.14     & 31.63     & 29.15    & 26.19    & 32.49 & 29.82 & 26.14   \\
   DnCNN    & 32.86   & 30.43   & 27.18     & 31.73     & 29.23    & 26.23    & 32.68 & 29.97 & 26.28   \\
   FFDNet    & 32.75   & 30.43   & 27.32     & 31.63     & 29.19    & 26.29    & 32.42 & 29.92 & 26.52  \\
   MLEFGN  & 33.04   & 30.66   & 27.54     & 31.81     & 29.34    & 26.39    & 33.21 & 30.64 & 27.22   \\
   MFENANN     & 32.95   & 30.63   & 27.55     & 31.73     & 29.29    & 26.38    & -     & -     &        \\
   DRNet      & 33.01   & 30.64   & 27.46     & 31.81     & 29.35    & 26.39    & -     & -     &        \\ \hline 
   \textbf{MDRN (Ours)}                     & \emph{33.06}  & \emph{30.67}   &   \emph{27.56}        & \emph{31.83}     & \emph{29.36}    &  \emph{26.41}        & \emph{33.22} & \emph{30.67} &  \emph{27.24}     \\
   \textbf{MDRN+ (Ours)}                    & \textbf{33.10}  & \textbf{30.71}   &   \textbf{27.60}        & \textbf{31.86}     & \textbf{29.39}    &   \textbf{26.44}       & \textbf{33.31} & \textbf{30.78} &    \textbf{27.31}   \\ \hline
   \end{tabular}
   \caption{PSNR comparison with classic SID methods on \textbf{grayscale} images with noise levels $\sigma=15, 25$, and $50$.}
   \label{gray-results}
\end{table*}

\begin{figure*}
\centering
\begin{minipage}[c]{0.135\textwidth}
\includegraphics[width=2.4cm,height=1.4cm]{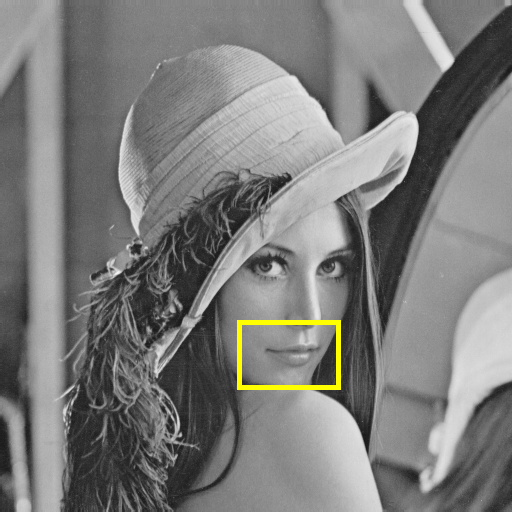}
\centerline{\small Lena}
\includegraphics[width=2.4cm,height=1.4cm]{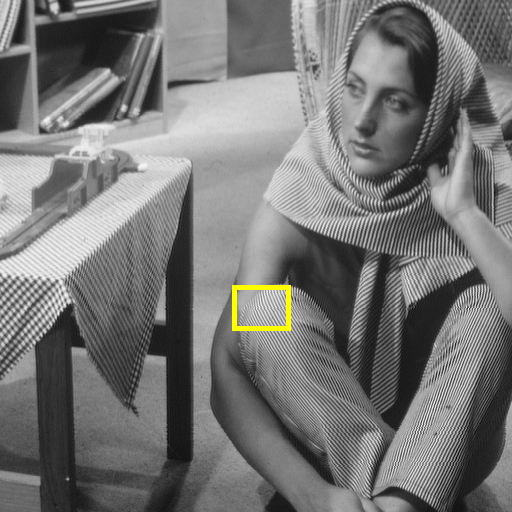}
\centerline{\small Barbara}
\centerline{\small }
\end{minipage}
\begin{minipage}[c]{0.135\textwidth}
\includegraphics[width=2.4cm,height=1.4cm]{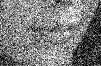}
\centerline{\small 14.61/0.1192}
\includegraphics[width=2.4cm,height=1.4cm]{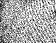}
\centerline{\small 14.77/0.2073}
\centerline{\small Noise}
\end{minipage}
\begin{minipage}[c]{0.135\textwidth}
\includegraphics[width=2.4cm,height=1.4cm]{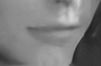}
\centerline{\small 29.39/0.8116}
\includegraphics[width=2.4cm,height=1.4cm]{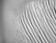}
\centerline{\small 26.22/0.7693}
\centerline{\small DnCNN}
\end{minipage}
\begin{minipage}[c]{0.135\textwidth}
\includegraphics[width=2.4cm,height=1.4cm]{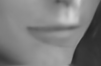}
\centerline{\small 29.68/0.8227}
\includegraphics[width=2.4cm,height=1.4cm]{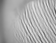}
\centerline{\small 26.48/0.7794}
\centerline{\small FFDNet}
\end{minipage}
\begin{minipage}[c]{0.135\textwidth}
\includegraphics[width=2.4cm,height=1.4cm]{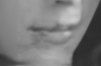}
\centerline{\small 29.85/0.8295}
\includegraphics[width=2.4cm,height=1.4cm]{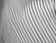}
\centerline{\small 27.24/0.8105}
\centerline{\small MLEFGN}
\end{minipage}
\begin{minipage}[c]{0.135\textwidth}
\includegraphics[width=2.4cm,height=1.4cm]{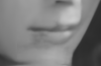}
\centerline{\small 29.87/0.8311}
\includegraphics[width=2.4cm,height=1.4cm]{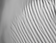}
\centerline{\small 27.37/0.8136}
\centerline{\small MDRN-Ours}
\end{minipage}
\begin{minipage}[c]{0.135\textwidth}
\includegraphics[width=2.4cm,height=1.4cm]{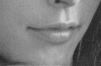}
\centerline{\small PSNR/SSIM}
\includegraphics[width=2.4cm,height=1.4cm]{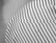}
\centerline{\small PSNR/SSIM}
\centerline{\small GT}
\end{minipage}

\caption{Visual comparison with DnCNN, FFDNet, and MLEFGN on grayscale images ($\sigma=50$). Zoom in to view details.}
\label{visual-grayscale}
\end{figure*}

\textbf{Knowledge Distillation Loss.} In this paper, we propose two heterogeneous knowledge distillation strategies.
In order to ensure that these two strategies can truly realize knowledge transfer, we propose a Knowledge Distillation Loss (KDL).
KDL is essentially a feature matching function, which can be defined as
\begin{equation}
   \mathcal{L}_{KDL} = \sum_{i=2,3,4} \left \| S^{i}_{G}(I_{noisy}) - T^{i}_{G}(I'_{noisy}) \right \| _{1},
\end{equation}
where $S^{i}_{G}(I_{noisy})$ and $T^{i}_{G}(I_{noisy})$ denote the output features of the $i^{th}$ MSAG in the student and teacher models, respectively.
Meanwhile, the noise level of the input noisy image $I_{noisy}=I'_{noisy}$ and $I_{noisy} \ne I'_{noisy}$ when using the HADS and HMDS, respectively. Therefore, the total loss is defined as
\begin{equation}
   \mathcal{L}_{total} = \mathcal{L}_{RL} + \mathcal{L}_{KDL}.
\end{equation}

\section{Experiment}
\subsection{Datasets}
Following previous works~\cite{zhang2017beyond, zhang2018ffdnet, fang2020multilevel, zhang2019residual, Yu2019DeepID}, we also choose the AWGN as our research object due to its extensiveness and practicality.
During training, we use DIV2K~\cite{agustsson2017ntire} as our training dataset since it is a high-quality image restoration dataset.
As for the test datasets, we use Set12~\cite{zeyde2010single}, BSD68~\cite{martin2001database}, and Urban100~\cite{huang2015single} for grayscale image denoising and choose Kodak24~\cite{FranzenKodak}, CBSD68~\cite{martin2001database}, and Urban100~\cite{huang2015single} for the color image denoising.
Meanwhile, in order to further verify the effectiveness and robustness of MDRN, we utilize RN6~\cite{Lebrun2015The} and
RN15~\cite{Lebrun2015The} to test the ability of MDRN for real noise removing. 

\subsection{Implementation Details}
\textbf{Model Setting.} In the final version of MDRN, the number of MSAB in each MSAG is set to $8$ and the input and output channels of each MSAB or MSAG are set to $64$.
In addition, the entire model contains three degradation processes, and the kernel size and stride are set to $2$ both in the convolutional and deconvolutional layers.
Meanwhile, we also introduce the self-ensemble mechanism~\cite{timofte2016seven} to further improve the model performance, which is denoted as MDRN+.
In addition, we provide a tiny version of MDRN, which contains 2 MSABs in MSAG, named MDRN (2B).

\textbf{Training Setting.} During training, we randomly choose 16 noisy patches as inputs, the learning rate is initialized as $10^{-4}$ and halved every $100$ epochs.
In addition, the MDRN is implemented with the PyTorch framework, updated with the Adam optimizer, and a total of $500$ epochs.
It is worth noting that when the heterogeneous knowledge distillation strategy is applied to MDRN to achieve joint training, the teacher model is pre-trained and the weights are fixed throughout the process without updating.

\subsection{Comparison with Classic SID Methods}
In order to verify the effectiveness of the proposed MDRN, we compare it with more than 10 classic SID methods, including BM3D~\cite{dabov2007image}, RED30~\cite{mao2016image}, TNRD~\cite{Chen2017Trainable}, IRCNN~\cite{zhang2017learning}, DnCNN~\cite{zhang2017beyond}, MemNet~\cite{tai2017memnet}, FFDNet~\cite{zhang2018ffdnet}, MLEFGN~\cite{fang2020multilevel}, MFENANN~\cite{wang2021multi}, and DRNe~\cite{zhang2021drnet}.
All aforementioned methods are the most widely used denoiser and all of them achieved the SOTA results at the time.

\begin{table*}
  \centering
  \small
  \setlength{\tabcolsep}{3.3mm}
  \renewcommand{\arraystretch}{0.8}
  \begin{tabular}{r|c|c|c|c|c|c|c|c|c}
  \hline
  Method    & \multicolumn{3}{c|}{Kodar24}    & \multicolumn{3}{c|}{CBSD68}     & \multicolumn{3}{c}{Urban100}  \\ \hline
  Noise Level    & $\sigma$=30      & $\sigma$=50      & $\sigma$=70     & $\sigma$=30       & $\sigma$=50      & $\sigma$=70       & $\sigma$=30    & $\sigma$=50    & $\sigma$=70    \\ \hline
  CBM3D       & 30.89   & 28.63   & 27.27     & 29.73    & 27.38   & 26.00    & 30.36 & 27.94 & 26.31\\
  RED30          & 29.71   & 27.62   & 26.36     & 28.46    & 26.35   & 25.08    & 29.02 & 26.40 & 24.74 \\
  TNRD    & 28.83   & 27.17   & 24.94     & 27.64    & 25.96   & 23.83    & 27.40 & 25.52 & 22.63 \\
  IRCNN   & 31.24   & 28.93   & N/A       & 30.22    & 27.86   & N/A      & 30.28 & 27.69 & N/A    \\
  DnCNN      & 31.39   & 29.16   & 27.64     & 30.40    & 28.01   & 26.56    & 30.28 & 28.16 & 26.17  \\
  MemNet  & 29.67   & 27.65   & 26.40     & 28.39    & 26.33   & 25.08    & 28.93 & 26.53 & 24.93  \\
  FFDNet    & 31.39   & 29.10   & 27.68     & 30.31    & 27.96   & 26.53    & 30.53 & 28.05 & 26.39 \\
  MLEFGN  & 31.67   & 29.38   & 27.94     &  30.56   & 28.21   & 26.75    & 31.32 & 28.92 & 27.28  \\ \hline 
  \textbf{MDRN (Ours)}                 &  \emph{31.68}   &    \emph{29.40}      & \emph{27.96}       &   \emph{30.57}       &   \emph{28.23}      & \emph{26.77}      &  \emph{31.35}   &    \emph{28.96}      & \emph{27.32}      \\
  \textbf{MDRN+ (Ours)}                &  \textbf{31.73}   &    \textbf{29.44}      & \textbf{28.01}       &   \textbf{30.61}        &   \textbf{28.27}      & \textbf{26.82}      &  \textbf{31.41}   &    \textbf{29.00}      & \textbf{27.37}       \\ \hline
  \end{tabular}
  \caption{PSNR comparison with classic SID methods on \textbf{color} images with noise levels $\sigma=30, 50$, and $70$.
  Best results are highlighted.}
  \label{color-results}
\end{table*}

\begin{figure*}
\centering
\begin{minipage}[c]{0.135\textwidth}
\includegraphics[width=2.4cm,height=1.4cm]{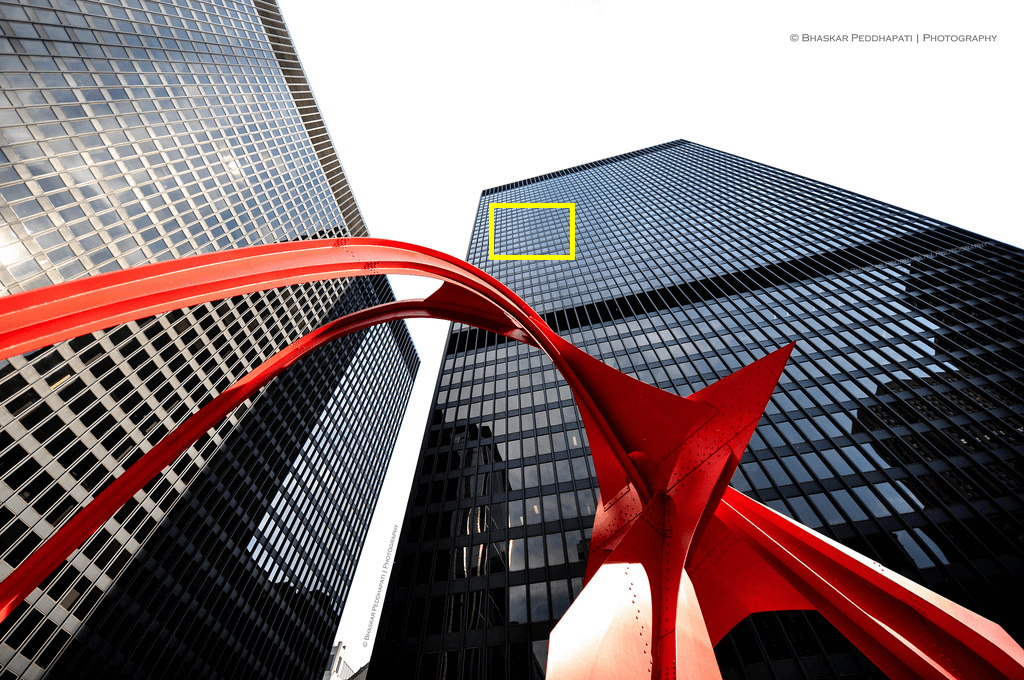}
\centerline{\small IMG062}
\includegraphics[width=2.4cm,height=1.4cm]{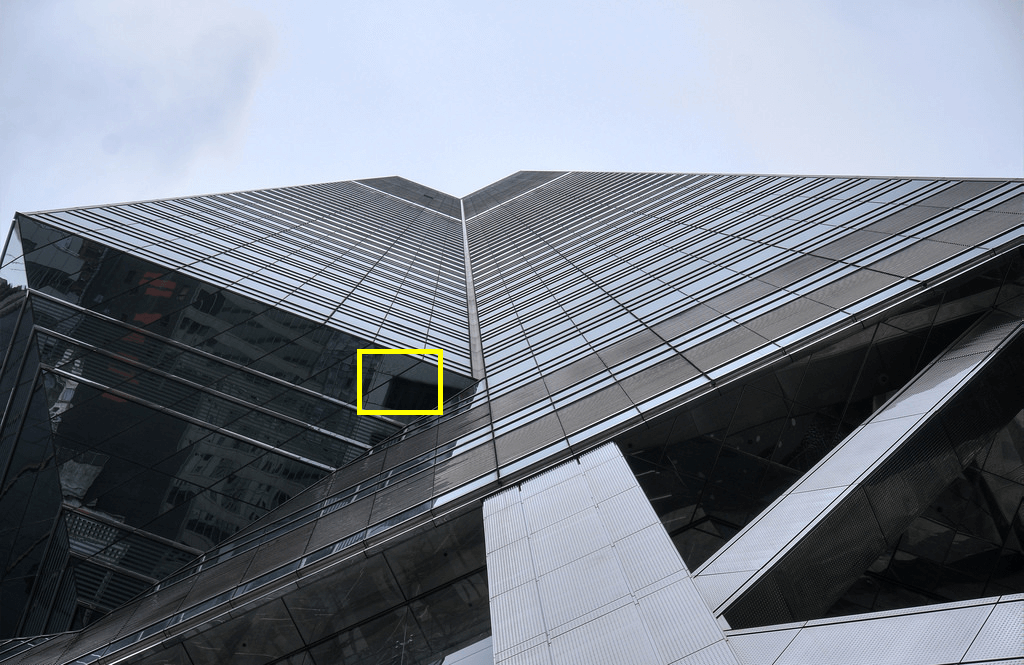}
\centerline{\small IMG059}
\centerline{\small }
\end{minipage}
\begin{minipage}[c]{0.135\textwidth}
\includegraphics[width=2.4cm,height=1.4cm]{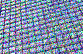}
\centerline{\small 15.91/0.3238}
\includegraphics[width=2.4cm,height=1.4cm]{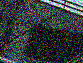}
\centerline{\small 15.62/0.2435}
\centerline{\small Noise}
\end{minipage}
\begin{minipage}[c]{0.135\textwidth}
\includegraphics[width=2.4cm,height=1.4cm]{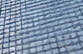}
\centerline{\small 27.14/0.8873}
\includegraphics[width=2.4cm,height=1.4cm]{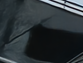}
\centerline{\small 29.26/0.8601}
\centerline{\small DnCNN}
\end{minipage}
\begin{minipage}[c]{0.135\textwidth}
\includegraphics[width=2.4cm,height=1.4cm]{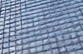}
\centerline{\small 27.04/0.8825}
\includegraphics[width=2.4cm,height=1.4cm]{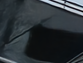}
\centerline{\small 29.10/0.8571}
\centerline{\small FFDNet}
\end{minipage}
\begin{minipage}[c]{0.135\textwidth}
\includegraphics[width=2.4cm,height=1.4cm]{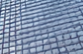}
\centerline{\small 27.73/0.8983}
\includegraphics[width=2.4cm,height=1.4cm]{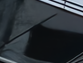}
\centerline{\small 30.04/0.8742}
\centerline{\small MLEFGN}
\end{minipage}
\begin{minipage}[c]{0.135\textwidth}
\includegraphics[width=2.4cm,height=1.4cm]{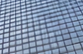}
\centerline{\small 28.28/0.9068}
\includegraphics[width=2.4cm,height=1.4cm]{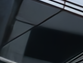}
\centerline{\small 30.67/0.8837}
\centerline{\small MDRN-Ours}
\end{minipage}
\begin{minipage}[c]{0.135\textwidth}
\includegraphics[width=2.4cm,height=1.4cm]{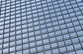}
\centerline{\small PSNR/SSIM}
\includegraphics[width=2.4cm,height=1.4cm]{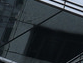}
\centerline{\small PSNR/SSIM}
\centerline{\small GT}
\end{minipage}

\caption{Visual comparison with DnCNN, FFDNet, and MLEFGN on color images ($\sigma=50$). Zoom in to view details.}
\label{visual-color}
\end{figure*}

\textbf{Results on Gray-Scale Images:} In Table~\ref{gray-results}, we provide the PSNR results of these methods on grayscale images.
According to the table, we can clearly observe that our MDRN+ and MDRN achieve the best and the second-best results, respectively.
In Figure~\ref{visual-grayscale}, we provide the visual comparison with DnCNN, FFDNet, and MLEFGN on grayscale images.
Among them, DnCNN and FFDNet achieve competitive results and still are the most widely used denoisers.
MLEFGN is a well-designed SID model, which achieved the best results under the same level parameters at the time.
According to the figure, we can observe that images reconstructed by these methods still contain a lot of noises and artifacts.
Moreover, denoised images become blurred since high-frequency details have been severely destroyed.
On the contrary, MDRN can reconstruct high-quality denoised images with more texture details.

\textbf{Results on Color Images:} To further verify the validity of the model, we also provide the PSNR results of these methods on color images in Table~\ref{color-results}.
According to the table, we can observe that MDRN still achieves the best results on color images.
Meanwhile, we also provide the denoised color images in Figure~\ref{visual-color}.
Obviously, MDRN can reconstruct high-quality denoised images with better image edges, even compared with MLEFGN.
All these results fully verified the effectiveness of the proposed MDRN.

\subsection{Results with HKDS} \label{HKDS}
Although our proposed MDRN has achieved excellent performance, it still does not work well under extreme conditions (tiny version models or severe noise levels).
In order to improve the denoising ability of the tiny model, we propose the Heterogeneous Architecture Distillation Strategy (HADS). 
Meanwhile, to improve the ability of the model to process images with high noise levels, we propose the Heterogeneous Mode Distillation Strategy (HMDS).
In this part, we show the model performance with these strategies.

\begin{table}
  \centering
  \small
  \setlength{\tabcolsep}{0.4mm}
  \renewcommand{\arraystretch}{0.86}
  \begin{tabular}{c|c|c|c|c|c|c|c}
  \hline
  Method & Para.   & \multicolumn{2}{c|}{Set12}    & \multicolumn{2}{c|}{BSD68}     & \multicolumn{2}{c}{Urban100}  \\ \hline
  Noise Level  & - & $\sigma$=15            & $\sigma$=50     & $\sigma$=15            & $\sigma$=50       & $\sigma$=15       & $\sigma$=50    \\ \hline
  FFDNet  & 851K   & 32.75     & 27.32     & 31.63        & 26.29    & 32.42  & 26.52 \\
  MDRN (2B)     & 770K       &    32.87           & 27.35     &   31.72             & 26.31    &  32.82         &  26.74     \\
  MDRN (2B) + HADS     & 770K      &  \textbf{32.96}         & \textbf{27.43}     &   \textbf{31.78}           &  \textbf{26.40}    &   \textbf{32.98}    &  \textbf{26.84}     \\  \hline
  \end{tabular}

  \caption{PSNR results of MDRN with and without the HADS on grayscale images. Best results are highlighted.}
  \label{HADS-results-1}
\end{table}

\begin{table}
  \centering
  \small
  \setlength{\tabcolsep}{0.4mm}
  \renewcommand{\arraystretch}{0.86}
  \begin{tabular}{c|c|c|c|c|c|c|c}
  \hline
  Method & Para.   & \multicolumn{2}{c|}{Kodak24}    & \multicolumn{2}{c|}{CBSD68}     & \multicolumn{2}{c}{Urban100}  \\ \hline
  Noise Level  & - & $\sigma$=30           & $\sigma$=70     & $\sigma$=30             & $\sigma$=70       & $\sigma$=30        & $\sigma$=70    \\ \hline
  FFDNet  & 851K   & 31.39     & 27.68     & 30.31       & 26.53    & 30.53  & 26.39 \\
  MDRN (2B)     & 770K       &    31.40          & 27.69     &   30.38            & 26.59    &  30.66         &  26.57     \\
  MDRN (2B) + HADS    & 770K      &  \textbf{31.47}              & \textbf{27.73}     &   \textbf{30.42}          &  \textbf{26.63}    &   \textbf{30.78}      &  \textbf{26.64}     \\  \hline
  \end{tabular}

  \caption{PSNR results of MDRN with and without the HADS on color images. Best results are highlighted.}
  \label{HADS-results-2}
\end{table}

\begin{figure}
\centering
\begin{minipage}[c]{0.11\textwidth}
\includegraphics[width=1.9cm, height=1.4cm]{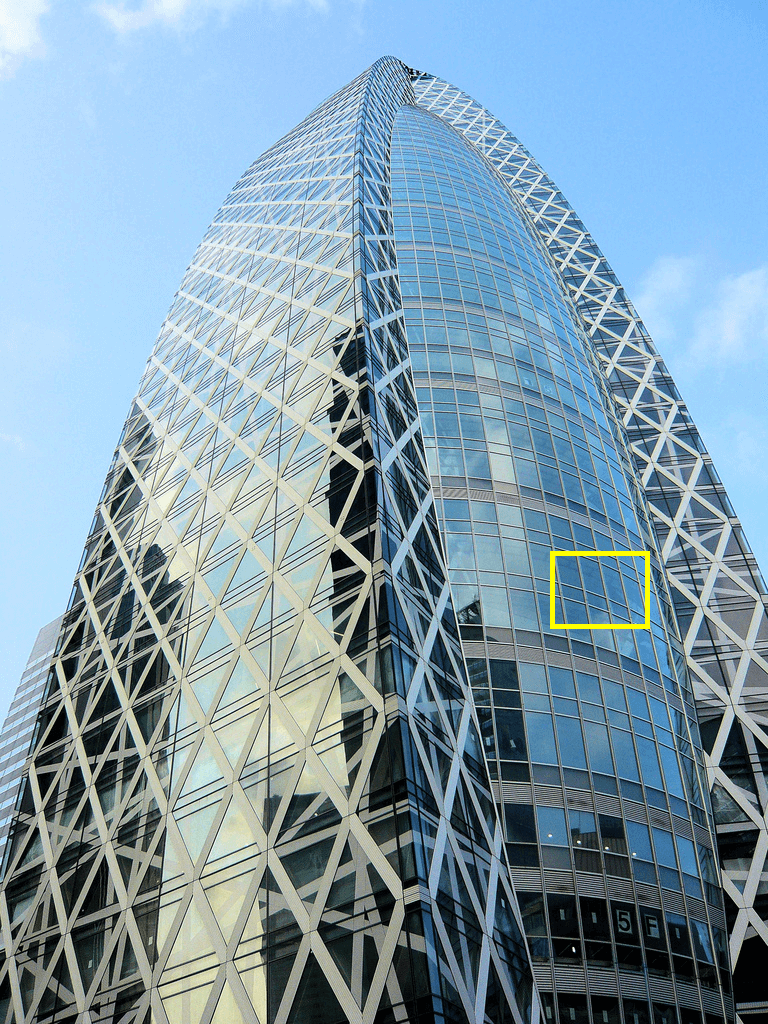}
\centerline{\small IMG039}
\includegraphics[width=1.9cm, height=1.4cm]{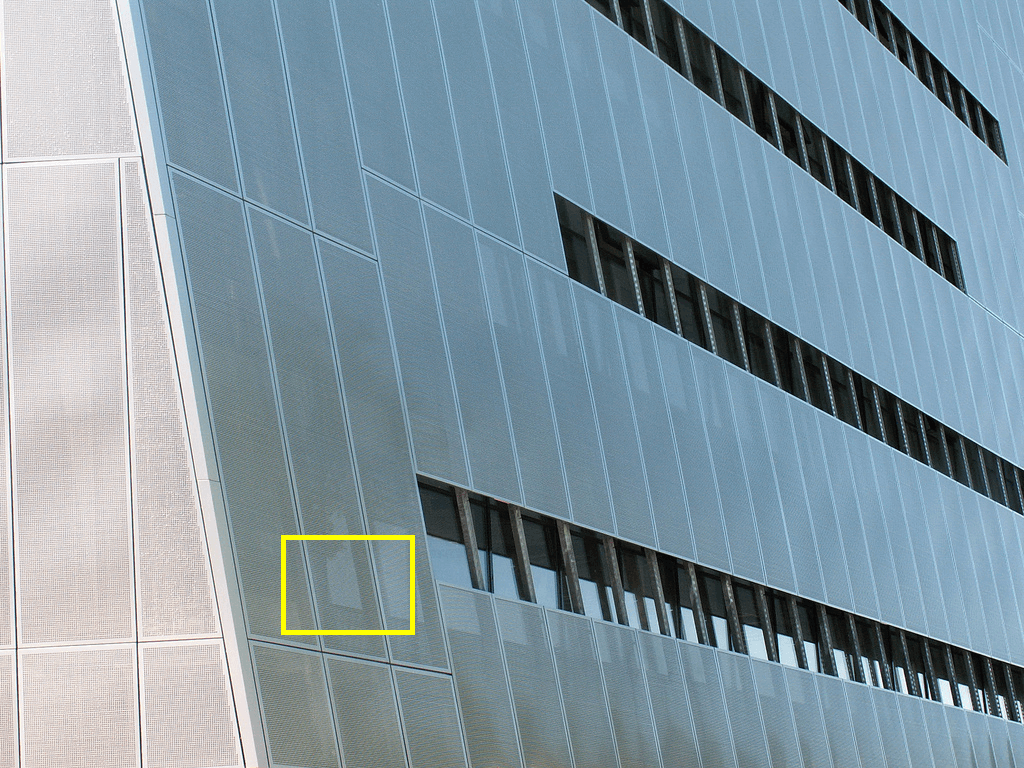}
\centerline{\small IMG026}
\end{minipage}
\begin{minipage}[c]{0.11\textwidth}
\includegraphics[width=1.9cm, height=1.4cm]{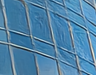}
\centerline{\small MDRN-2B}
\includegraphics[width=1.9cm, height=1.4cm]{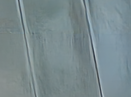}
\centerline{\small MDRN-2B}
\end{minipage}
\begin{minipage}[c]{0.11\textwidth}
\includegraphics[width=1.9cm, height=1.4cm]{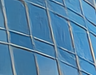}
\centerline{\small with HADS}
\includegraphics[width=1.9cm, height=1.4cm]{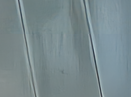}
\centerline{\small with HADS}
\end{minipage}
\begin{minipage}[c]{0.11\textwidth}
\includegraphics[width=1.9cm, height=1.4cm]{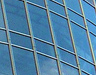}
\centerline{\small GT}
\includegraphics[width=1.9cm, height=1.4cm]{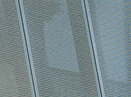}
\centerline{\small GT}
\end{minipage}

\begin{minipage}[c]{1\textwidth}
\end{minipage}
\caption{Visual comparison of MDRN (2B) with and without HADS on color images ($\sigma=70$). Zoom in to view details.}
\label{HADS-Visual}
\end{figure}

\textbf{Results with HADS.} 
The core idea of HADS is to use a large model as Teacher to guide the tiny model to learn more accurate features, thus further improving the quality of the reconstructed images.
In Tables~\ref{HADS-results-1} and ~\ref{HADS-results-2}, we show the PSNR results of the FFDNet and the tiny version of MDRN (2B) with and without HADS on grayscale and color images, respectively.
Compared to FFDNet, MDRN (2B) can achieve better results with fewer parameters.
However, we also found that this improvement is not significant.
To further improve the model performance, we used the pre-trained final MDRN ($N=8$) as the Teacher and introduced HADS during training.
According to these results, we can clearly observe that with the help of HADS, the performance of MDRN (2B) can be further improved.
Therefore, the new results are significantly improved compared to FFDNet.
In Figure~\ref{HADS-Visual}, we also provide the visual comparison of MDRN (2B) with and without HADS on color images.
Obviously, with the help of HADS, MDRN can reconstruct cleaner images with more accurate edges.
All the above experiments show that the proposed HADS is effective, which can further improve the lightweight model performance.

\begin{figure}
\centering
\begin{minipage}[c]{0.11\textwidth}
\includegraphics[width=1.9cm, height=1.4cm]{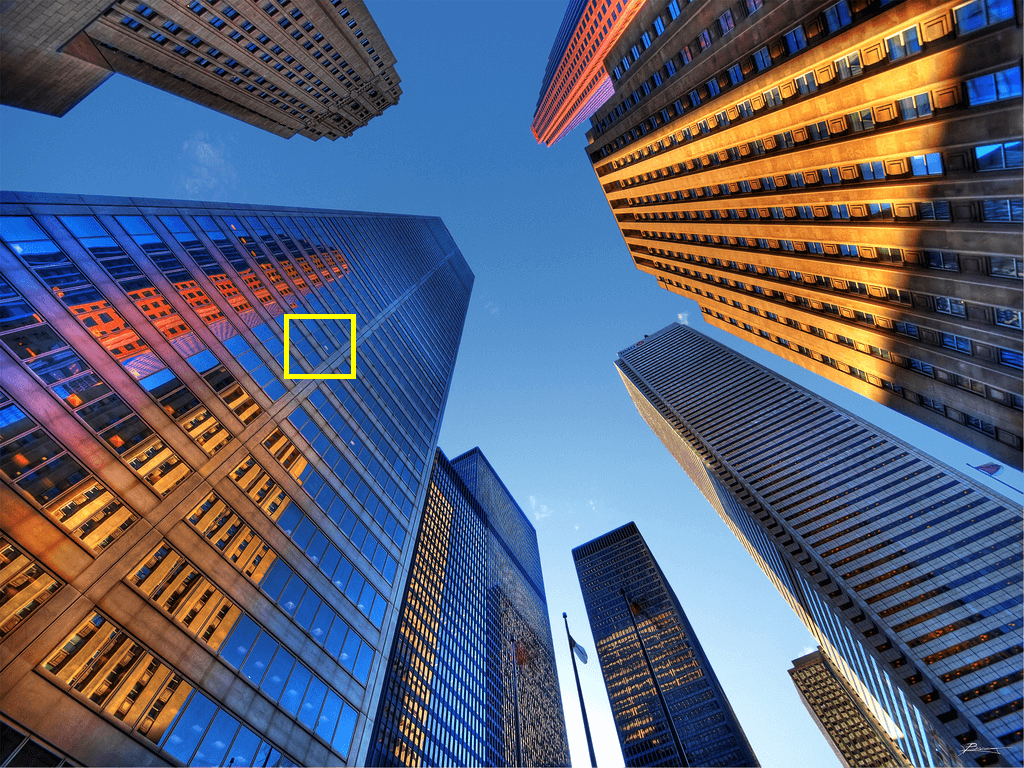}
\centerline{\small IMG012}
\includegraphics[width=1.9cm, height=1.4cm]{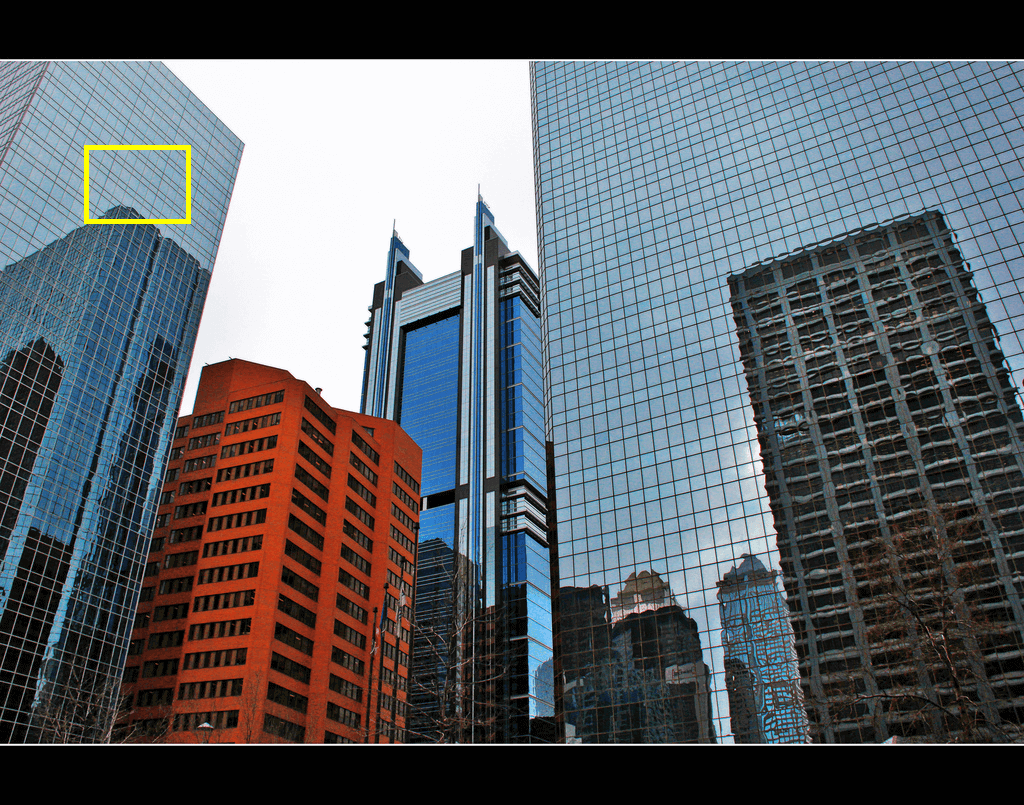}
\centerline{\small IMG099}
\end{minipage}
\begin{minipage}[c]{0.11\textwidth}
\includegraphics[width=1.9cm, height=1.4cm]{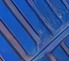}
\centerline{\small MDRN}
\includegraphics[width=1.9cm, height=1.4cm]{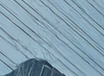}
\centerline{\small MDRN}
\end{minipage}
\begin{minipage}[c]{0.11\textwidth}
\includegraphics[width=1.9cm, height=1.4cm]{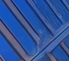}
\centerline{\small with HMDS}
\includegraphics[width=1.9cm, height=1.4cm]{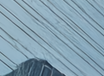}
\centerline{\small with HMDS}
\end{minipage}
\begin{minipage}[c]{0.11\textwidth}
\includegraphics[width=1.9cm, height=1.4cm]{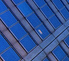}
\centerline{\small GT}
\includegraphics[width=1.9cm, height=1.4cm]{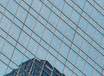}
\centerline{\small GT}
\end{minipage}

\caption{Visual comparison of MDRN with and without HMDS (50) on color images ($\sigma=70$). Zoom in to view details.}
\label{HMDS-Visual}
\end{figure}

\textbf{Results with HMDS.} High noise level image denoising is still a challenging task since noise will severely damage the image.
Recently, the most common method to solve this problem is to build larger CNN models.
Different from previous works, we propose HMDS to enable the model to learn more accurate high-frequency details from the Teacher.
Specifically, according to the core idea of the strategy, we use the model trained under $\sigma=25$ or $\sigma=50$ as the teacher model to guide the model trained for $\sigma=50$ or $\sigma=70$.
In Tables~\ref{HMDS-results}, we show the PSNR results of MDRN with and without HMDS.
According to the results, we can observe that (1). compared to MLEFGN~\cite{fang2020multilevel}, MDRN can achieve better results with fewer parameters; (2). with the help of HMDS, the performance of our MDRN can be further improved.
Similarly, we also provide the visual comparison of MDRN with and without HMDS on color images in Figure~\ref{HMDS-Visual}.
Obviously, when the image suffers from severe noise interference, it is extremely difficult to restore clean images.
Fortunately, with the help of HMDS, MDRN can reconstruct more clean and accurate denoised images.
All aforementioned experiments fully demonstrate the effectiveness of the proposed HMDS.

\begin{table}
  \centering
  \small
\setlength{\tabcolsep}{0.7mm}
  \begin{tabular}{c|c|c|c|c}
  \hline
  Method    & Parm.    & Set12      & BSD68         & Urban100    \\ \hline 
  Noise Level  &  -  & $\sigma$=50      & $\sigma$=50            & $\sigma$=50    \\ \hline 
  MLEFGN  &  6.86M & 27.54      & 26.39     & 27.22 \\
  MDRN          & 2.38M               & 27.56      & 26.41     & 27.24      \\
  MDRN + HMDS (25)  & 2.38M   & \textbf{27.63}   & \textbf{26.46}   &   \textbf{27.38}    \\  \hline \hline

  Method    & Parm.    & Kodak24      & CBSD68         & Urban100    \\ \hline 
  Noise Level  &  -  & $\sigma$=70      & $\sigma$=70            & $\sigma$=70    \\ \hline 
  MLEFGN  &  6.86M & 27.94      & 26.75     & 27.28 \\
  MDRN          & 2.38M               & 27.96      & 26.77     & 27.32      \\
  MDRN + HMDS (50)  & 2.38M   & \textbf{28.00}   & \textbf{26.81}   &   \textbf{27.45}    \\  \hline
  \end{tabular}

  \caption{PSNR results of MDRN with and without HMDS on grayscale and color images, respectively.}
  \label{HMDS-results}
\end{table}

\subsection{Study on High Noise Level}
\begin{figure}
  \centering

  \begin{minipage}[c]{0.11\textwidth}
  \includegraphics[width=1.9cm,height=1.4cm]{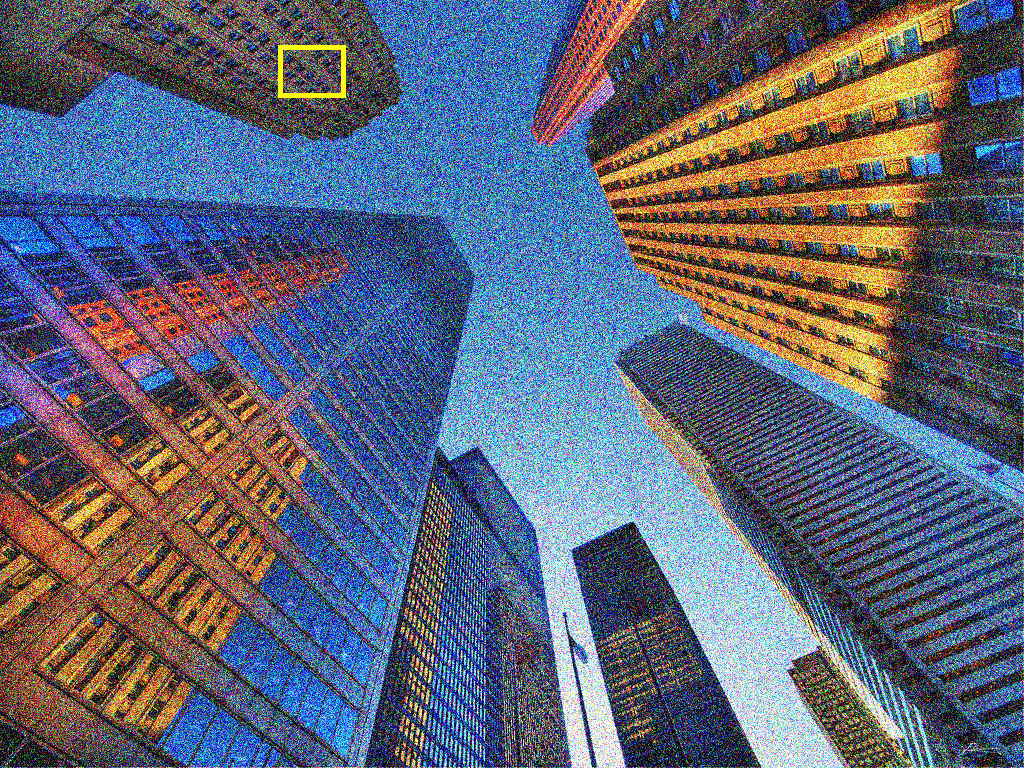}
  \includegraphics[width=1.9cm,height=1.4cm]{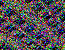}
  \centerline{Noisy image}
  \end{minipage}
  \begin{minipage}[c]{0.11\textwidth}
  \includegraphics[width=1.9cm,height=1.4cm]{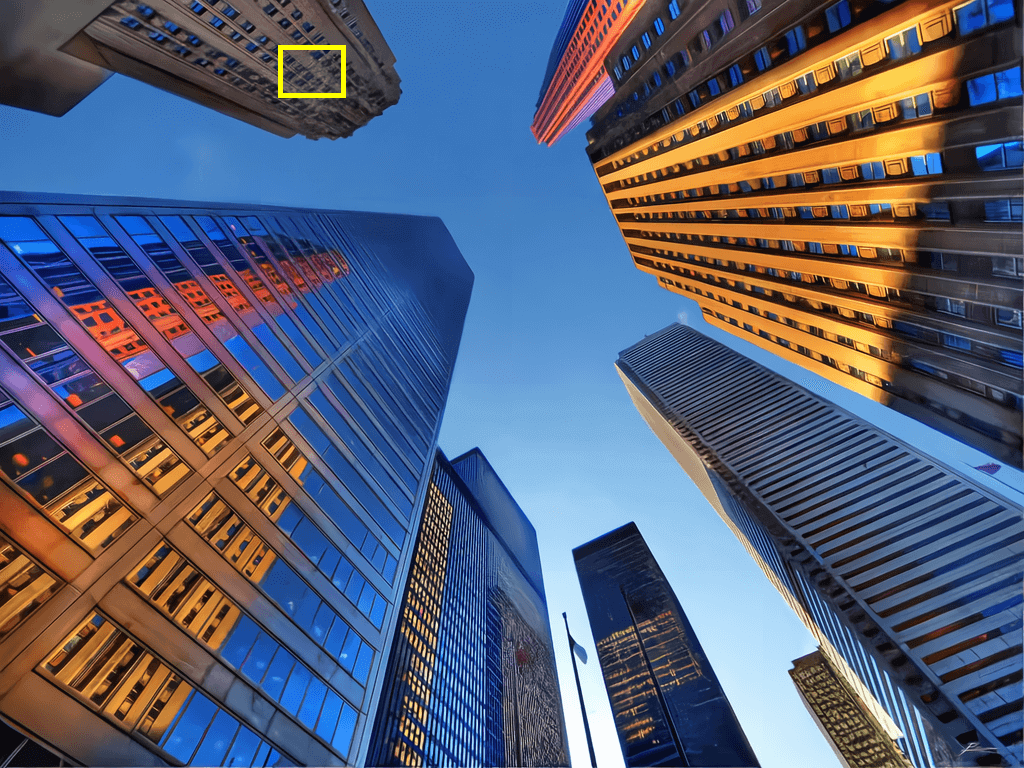}
  \includegraphics[width=1.9cm,height=1.4cm]{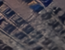}
  \centerline{MLEFGN}
  \end{minipage}
  \begin{minipage}[c]{0.11\textwidth}
  \includegraphics[width=1.9cm,height=1.4cm]{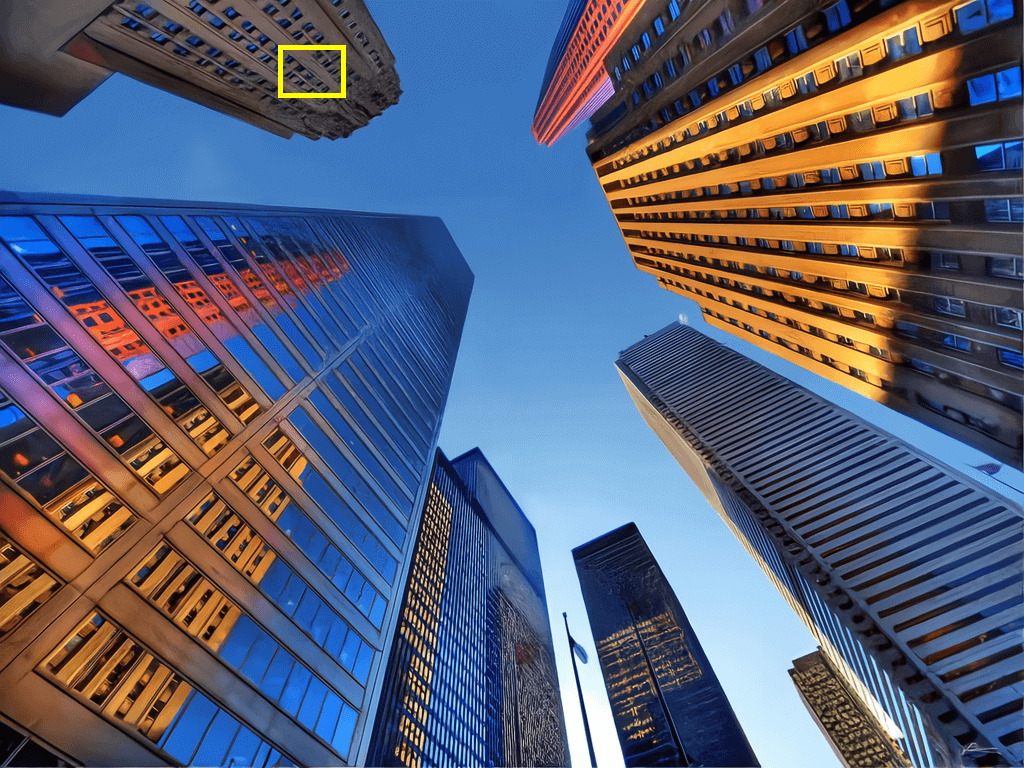}
  \includegraphics[width=1.9cm,height=1.4cm]{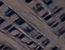}
  \centerline{MDRN-Ours}
  \end{minipage}
  \begin{minipage}[c]{0.11\textwidth}
  \includegraphics[width=1.9cm,height=1.4cm]{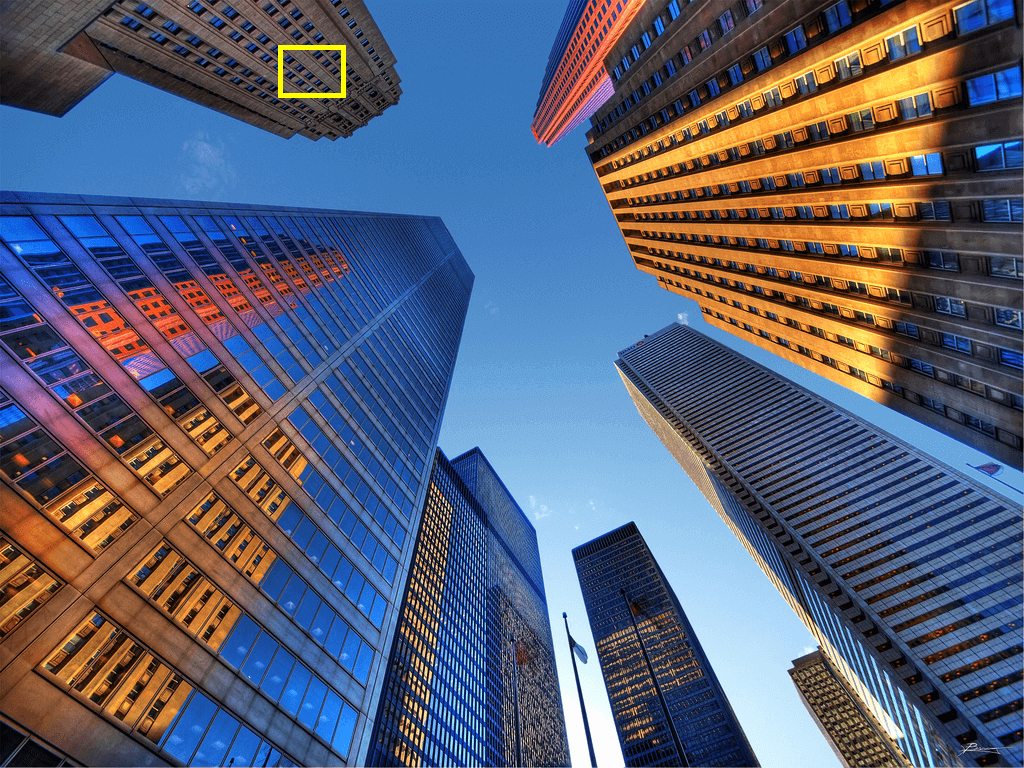}
  \includegraphics[width=1.9cm,height=1.4cm]{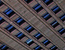}
  \centerline{GT}
  \end{minipage}
  
  \caption{Visual comparison of the denoising results of MDRN and MLEFGN on high-noise images (noise level: $\sigma=70$).}
  \label{visual-3}
\end{figure}

High noise level image denoising is still a challenging task since noise will severely damage images. 
In this part, we provide more denoising results of high-noise images in Figure~\ref{visual-3}.
According to the results, we can observe that MLEFGN is still difficult to reconstruct high-quality denoised images under the high noise level even though it is an excellent SID model.
On the contrary, our MDRN can reconstruct clearer and more accurate denoised images, even the model parameters of MDRN is only one-third of MLEFGN.
This is because MDRN can extract rich multi-scale image features, thus it can reconstruct more accurate denoised images.

\subsection{Study on Model Size}\label{size}
\begin{figure}
  \centering
  \includegraphics[width=7.6cm]{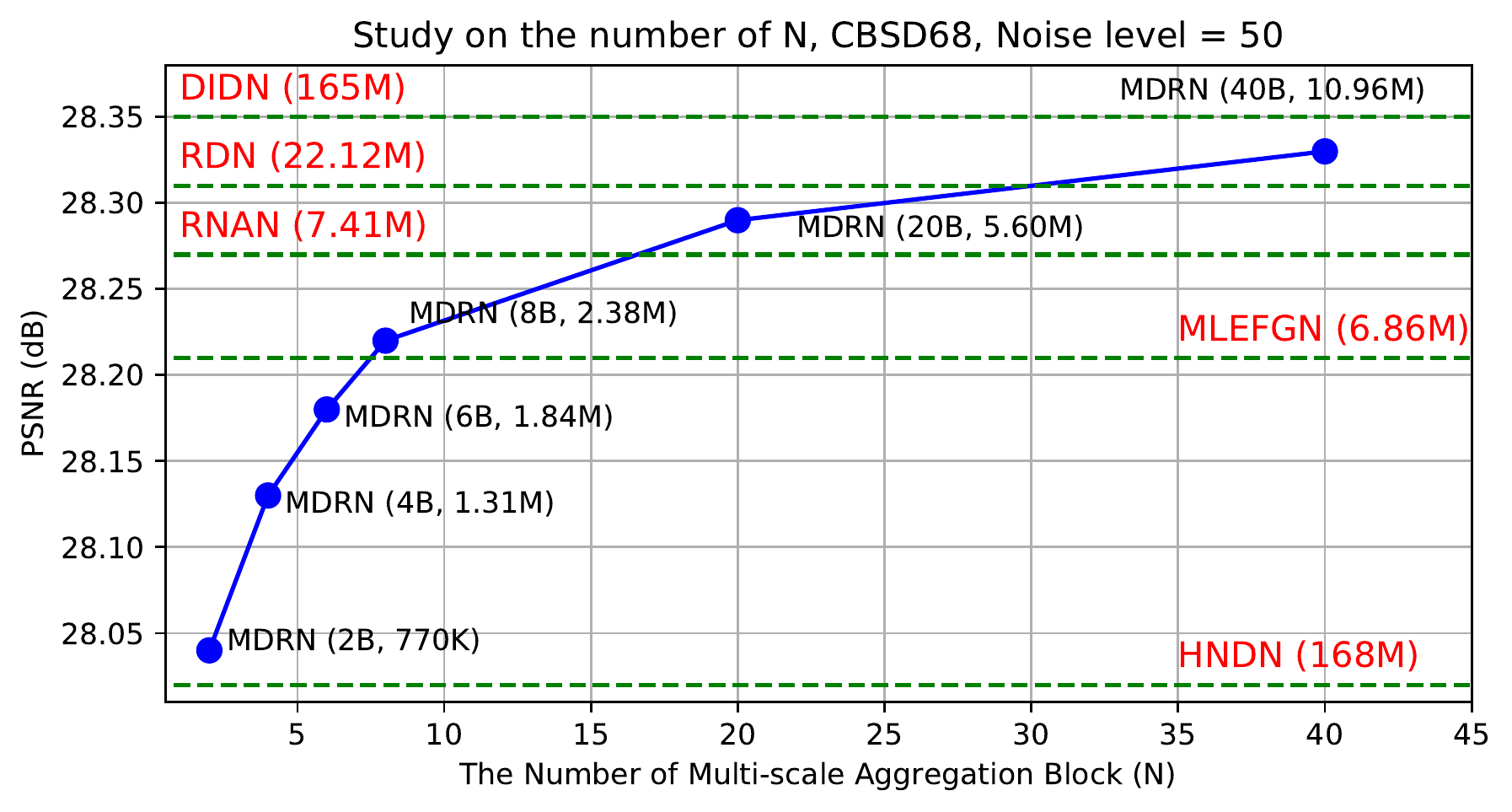}
  \caption{Study on the number ($N$) of MSAB in each MSAG.}
  \label{Block}
\end{figure}

Recently, some outstanding SID models have been proposed, such as RDN~\cite{zhang2020residual}, RNAN~\cite{zhang2019residual}, DIDN~\cite{Yu2019DeepID}, and DHDN~\cite{Park2019DenselyCH}.
However, it cannot be ignored that the parameters of these models are ten or even hundred times that of our MDRN.
In this paper, we provide a lightweight MDRN, which achieves a good balance between the size and performance of the model.
However, we should notice that the model size of MDRN can be easily changed by changing the number of MSAB ($N$) in MSAG.
In Figure~\ref{Block}, we provide the model performance with different $N$. According to the figure, we can find (1). as $N$ increases, the model performance can be further improved; (2). when $N=20$, MDRN exceeds RNAN with fewer parameters; (3). when $N=40$, MDRN exceeds RDN with only half of the parameters. Meanwhile, MDRN achieves close performance to DIDN with only one-fifteenth of the parameters. This fully demonstrates the effectiveness and potential of MDRN.

Moreover, we provide a more intuitive comparison in Figure~\ref{parameter}.
It is worth noting that the point in the upper left corner of the figure represents a better balance between model size and performance.
According to the figure, we can clearly observe that our MDRN achieves the best tradeoff between model size and performance.

\section{Conclusion}
In this paper, we proposed a lightweight and efficient Multiple Degradation and Reconstruction Network (MDRN) for SID, which realizes automatic noise removal and feature restoration through multiple down- and up-sampling operations.
Meanwhile, we proposed two novel strategies, namely Heterogeneous Architecture Distillation Strategy (HADS) and Heterogeneous Mode Distillation Strategy (HMDS) to improve the performance of tiny models and the model under high noise levels, respectively.
Extensive experiments have fully proved the effectiveness of the proposed model and strategies.

\section{Acknowledgment}
This work was supported in part by the National Key R\&D Program of China under Grant 2021YFE0203700, Grant NSFC/RGC N\_CUHK 415/19, Grant ITF MHP/038/20, Grant RGC 14300219, Grant CRF 8730063, in part by the National Natural Science Foundation of China under Grant 61871185, and in part by the Shanghai Rising-Star Program under Grant 21QA1402500.

{\small
\bibliographystyle{ieee_fullname}
\bibliography{egbib}
}

\end{document}